\documentclass[letterpaper, 10 pt, conference]{ieeeconf} 

\IEEEoverridecommandlockouts
\usepackage{cite}
\usepackage{amsmath,amssymb,amsfonts}
\usepackage{algorithmic}
\usepackage{graphicx}
\usepackage[caption=false,font=footnotesize]{subfig}
\usepackage{multirow} 
\usepackage{textcomp}
\usepackage{tabularray}
\usepackage{array}
\usepackage{soul,color}
\usepackage[table]{xcolor}
\usepackage[inkscapeformat=pdf]{svg}
\usepackage{booktabs}
\def\BibTeX{{\rm B\kern-.05em{\sc i\kern-.025em b}\kern-.08em
    T\kern-.1667em\lower.7ex\hbox{E}\kern-.125emX}}

\usepackage{tikz} 
\usepackage{float} 
\usepackage{stfloats}
\usepackage[utf8]{inputenc} 
\usepackage{siunitx}
\usepackage{booktabs} 
\usepackage{amsmath}
\usepackage[ruled,vlined]{algorithm2e}
\usepackage[nolist]{acronym} 
\usepackage{threeparttable}
\usepackage[percent]{overpic}
\usepackage{refcount}
\usepackage{hyperref}

\def\BibTeX{{\rm B\kern-.05em{\sc i\kern-.025em b}\kern-.08em
    T\kern-.1667em\lower.7ex\hbox{E}\kern-.125emX}}

\usepackage{scalerel} 
\usepackage{tikz} 
\usetikzlibrary{svg.path} 

\definecolor{poscolor}{HTML}{3D405B}
\definecolor{orcidlogocol}{HTML}{A6CE39}
\tikzset{
  orcidlogo/.pic={
    \fill[orcidlogocol] svg{M256,128c0,70.7-57.3,128-128,128C57.3,256,0,198.7,0,128C0,57.3,57.3,0,128,0C198.7,0,256,57.3,256,128z};
    \fill[white] svg{M86.3,186.2H70.9V79.1h15.4v48.4V186.2z}
                 svg{M108.9,79.1h41.6c39.6,0,57,28.3,57,53.6c0,27.5-21.5,53.6-56.8,53.6h-41.8V79.1z M124.3,172.4h24.5c34.9,0,42.9-26.5,42.9-39.7c0-21.5-13.7-39.7-43.7-39.7h-23.7V172.4z}
                 svg{M88.7,56.8c0,5.5-4.5,10.1-10.1,10.1c-5.6,0-10.1-4.6-10.1-10.1c0-5.6,4.5-10.1,10.1-10.1C84.2,46.7,88.7,51.3,88.7,56.8z};
  }
}

\newcommand\orcidicon[1]{\href{https://orcid.org/#1}{\mbox{\scalerel*{
\begin{tikzpicture}[yscale=-1,transform shape]
\pic{orcidlogo};
\end{tikzpicture}
}{|}}}}

\begin{document}

\title{\LARGE \bf 
Thermal Image Refinement with Depth Estimation using \\Recurrent Networks for Monocular ORB-SLAM3}

%

\author{Hürkan Şahin$^{1}$\textsuperscript{\orcidicon{0009-0008-7920-5872}}, Huy Xuan Pham$^{2}$\textsuperscript{\orcidicon{0000-0001-8218-9326}}, Van Huyen Dang$^{1}$\textsuperscript{\orcidicon{0009-0006-2328-2153}}, Alper Yegenoglu$^{1}$\textsuperscript{\orcidicon{0000-0001-8869-215X}}, and Erdal Kayacan$^{1}$\textsuperscript{\orcidicon{0000-0002-7143-8777}}
\thanks{*This work was partially supported by the Horizon Europe Grant Agreement No. 101136056 and No. 101070405, and Independent Research Fund Denmark, DFF-Research Project 1, with case number: 2035-00052B.}
\thanks{$^{1}$Hürkan Şahin, Van Huyen Dang, Alper Yegenoglu, and Erdal Kayacan are with the Automatic Control Group (RAT), Paderborn University, 33098 Paderborn, Germany
        {\tt\footnotesize \{hursah, van.huyen.dang, alper.yegenoglu, erdal.kayacan\}@upb.de}}%
\thanks{$^{2}$Huy Xuan Pham is with the Department of Electrical and Computer Engineering, Aarhus University, 8000 Aarhus C, Denmark, and also with Upteko ApS, Denmark {\tt\footnotesize huy.xuan@upteko.com}} 
\thanks{
$^\dagger$ Our dataset and source code are publicly available at 
\footnotesize \url{https://hurkansah.github.io/thermal-depth-orbslam3/}.       
       }
       }

\maketitle

\begin{abstract}
Autonomous navigation in GPS-denied and visually degraded environments remains challenging for \acp{UAV}. To this end, we investigate the use of a monocular thermal camera as a standalone sensor on a UAV platform for real-time depth estimation and \ac{SLAM}. To extract depth information from thermal images, we propose a novel pipeline employing a lightweight supervised network with \acp{RB} integrated to capture temporal dependencies, enabling more robust predictions. The network combines lightweight convolutional backbones with a \ac{tref} to refine raw thermal inputs and enhance feature visibility. The refined thermal images and predicted depth maps are integrated into ORB-SLAM3, enabling thermal-only localization. Unlike previous methods, the network is trained on a custom non-radiometric dataset, obviating the need for high-cost radiometric thermal cameras. Experimental results on datasets and UAV flights demonstrate competitive depth accuracy and robust SLAM performance under low-light conditions. On the radiometric VIVID++ (indoor–dark) dataset, our method achieves an absolute relative error of approximately 0.06, compared to baselines exceeding 0.11. In our non-radiometric indoor set, baseline errors remain above 0.24, whereas our approach remains below 0.10. Thermal-only ORB-SLAM3 maintains a mean trajectory error under 0.4 m.
\end{abstract}
\begin{acronym}
  \acro{UAV}[UAV]{unmanned aerial vehicle}
  \acro{CNN}[CNN]{convolutional neural network}
  \acro{SLAM}[SLAM]{simultaneous localization and mapping}
  \acro{HDR}[HDR]{high dynamic range}
  \acro{RB}[RB]{recurrent block}
  \acro{tref}[T-RefNet]{thermal refinement network}
  \acro{lif}[LIF]{leaky-integrate and fire}
  \acro{RC}[RC]{reservoir computing}
  \acro{TBB}[TBB]{target blackbody}
  \acro{NUC}[NUC]{non-uniformity correction}
  \acro{AGC}[AGC]{automatic gain control}
  \acro{CLAHE}[CLAHE]{contrast limited adaptive histogram equalization}
\end{acronym}

\section{Introduction}




In recent decades, research on autonomous \ac{UAV}s has accelerated, broadening their range of applications, including environmental monitoring \cite{10155874, 9838352}, infrastructure inspection \cite{10406329, 11010821}, and disaster response \cite{9352080}. In search and rescue missions, rescue robots and \ac{UAV}s can rapidly access hazardous or confined spaces, reduce risks to responders, and use advanced sensors for monitoring and localization, thereby enhancing situational awareness and mission success \cite{rescuedrone}. Recent advances in autonomous aerial navigation favor lightweight, low-cost cameras over LiDAR, yet maintaining robustness in degraded or dark conditions remains a challenge.
Thermal-infrared cameras thus provide key advantages under degraded conditions~\cite{thermalslam}, as their working principle allows detecting infrared radiation without requiring light exposure, enabling penetration through smoke, dust, or haze. 

Although beneficial under degraded conditions, thermal cameras pose distinct challenges for reliable \ac{SLAM} integration\cite{9804793}. The 14/16-bit high dynamic range of thermal camera conflicts with 8-bit vision algorithms, \ac{AGC} causes temporal inconsistencies, \ac{NUC} interrupts streams, and low texture hampers feature detection. These factors necessitate specialized adaptation of thermal imagery for robust \ac{SLAM}. 
\begin{figure}[!t]
    \centering
    \includegraphics[width=0.47\textwidth]{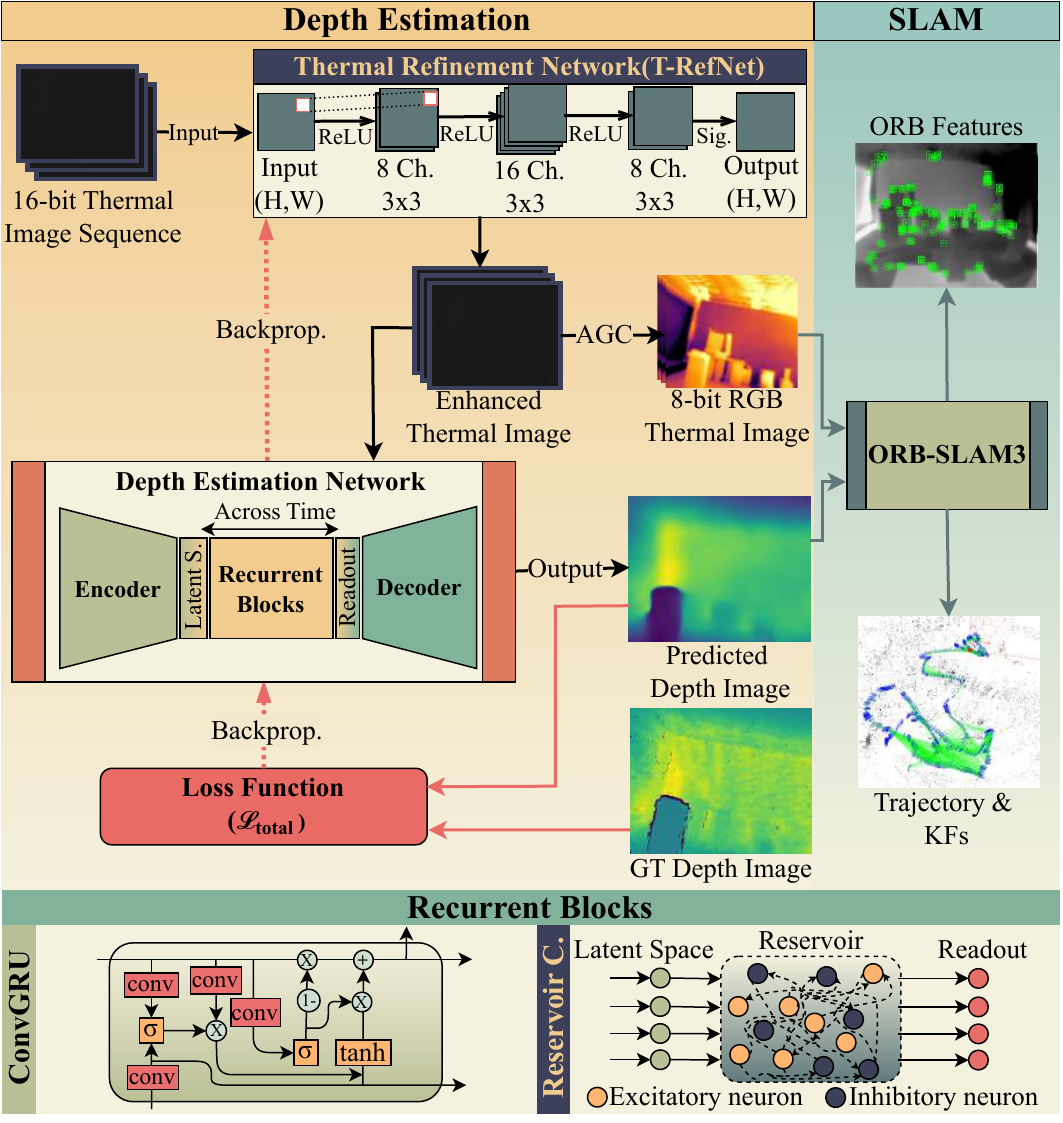}
    \caption{Overview of the proposed thermal depth estimation pipeline.
A raw 16-bit long-wave infrared (LWIR) image is first enhanced by the \ac{tref} module, producing both an enhanced input for depth prediction and a color-mapped image for robust ORB-SLAM3 feature extraction. The encoder backbone extracts multi-scale features, which are processed by \acp{RB} (ConvGRU~\cite{convgru} or \acf{RC}~\cite{jaeger2001echo}) to enforce temporal consistency. Finally, the decoder outputs dense depth maps and enhanced thermal images integrated into ORB-SLAM3~\cite{orbslam3} for robust feature extraction and metric-scale, temporally consistent tracking.}
    \label{fig:method_pipeline}
    \vspace{-1.5mm}
\end{figure}
To handle these challenges, we propose a novel framework (Fig.~\ref{fig:method_pipeline}) that leverages recurrent thermal-to-depth modeling for monocular depth estimation from thermal imagery. The enhanced thermal images and reconstructed depth maps provide metric scale and temporally consistent priors that can be directly integrated into ORB-SLAM3, improving initialization, mapping accuracy, and real-time tracking for autonomous \ac{UAV} navigation under extreme conditions.

\begin{figure*}[b]
    \centering
    \begin{overpic}[width=0.98\textwidth]{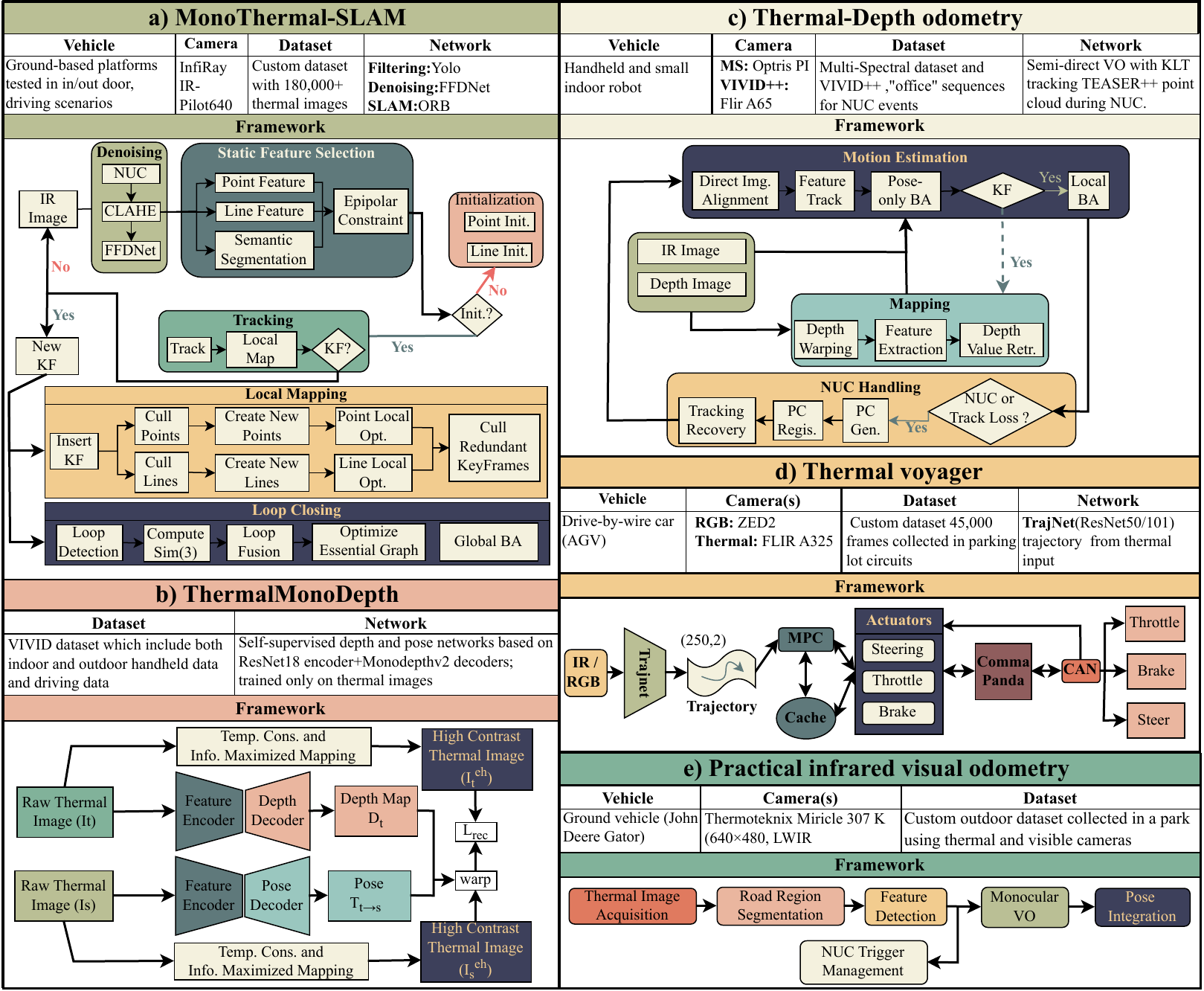}
        \put(34,80.35){\small \textbf{\cite{WU2023265}}}       
        \put(33.5,32.3){\small \textbf{\cite{ShinMaximizing}}} 
        \put(85,80.3){\small \textbf{\cite{10111061}}}       
        \put(82,42.5){\small \textbf{\cite{10611311}}}     
        \put(88.8,17.8){\small \textbf{\cite{7399731}}}        
    \end{overpic}
    \caption{Recent thermal navigation frameworks found in the literature span ground, handheld, and indoor platforms, across datasets from urban driving to parking-lot and outdoor road scenes. Representative approaches include feature/semantics-aware tracking and point–line SLAM~[11], self-supervised depth–ego-motion~[12], NUC handling~[11,13,15], LWIR-based trajectory prediction with MPC~[14], and road-segmentation–based scale recovery~[15].}
    \label{fignav:thermal_nav}
\end{figure*}
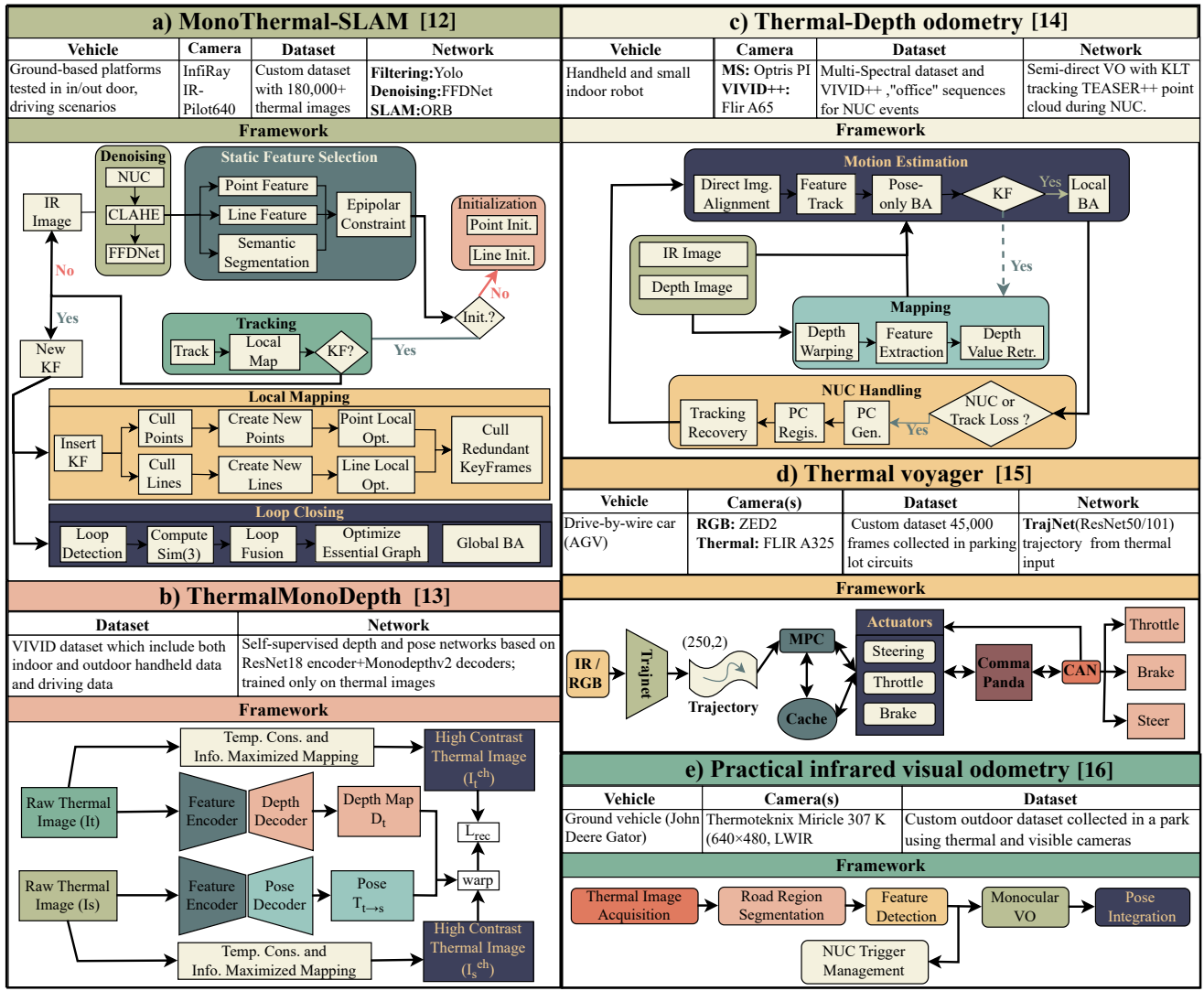
    
    

    The key contributions of this paper are as follows:

\begin{itemize}
\item We propose a lightweight framework, T-RefNet, that leverages a recurrent unit to enhance thermal-to-depth conversion. Our framework can use ResNet\cite{he2015deepresiduallearningimage}, EfficientNet\cite{tan2020efficientnetrethinkingmodelscaling}, and MobileNet\cite{howard2017mobilenetsefficientconvolutionalneural} to serve as a backbone, combined with recurrent architectures: ConvGRU\cite{ma2020convgrufinegrainedpitchingaction} and \acf{RC}~\cite{jaeger2001echo} to improve feature visibility and enforce temporal consistency in low-contrast and non-radiometric thermal imagery. 
    
\item We propose a non-radiometric thermal–depth~\ac{UAV} dataset to evaluate our framework, alongside existing radiometric public datasets such as VIVID++~\cite{vivid_dataset}.
    
    \item A comprehensive experimental study, including real-world experiments, is conducted to illustrate reliable performance in a thermal-only robot localization task across diverse trajectories and illumination settings, including fully dark environments where RGB-based \ac{SLAM} typically fails.
\end{itemize}

The remainder of this paper is organized as follows. Section~\ref{sec:relatedwork} summarizes related work, while Section~\ref{sec:methodology} provides a brief overview of our methodology. Section~\ref{sec:experiment} details the experimental setup, dataset, and real-time results. Finally, conclusions are presented in Section~\ref{sec:conclusion}.

\section{Related work}\label{sec:relatedwork}
 
The literature presents a wide range of thermal-based SLAM and visual odometry methods. As illustrated in Fig.~\ref{fignav:thermal_nav}, these approaches adopt diverse strategies to overcome the inherent challenges of thermal imaging, thereby enabling robust navigation in GPS-denied and visually degraded environments.

 A feature-based monocular SLAM framework is proposed in \cite{WU2023265} to address challenges in dynamic and visually degraded environments, combining thermal image denoising, semantic segmentation, and hybrid point-line tracking to improve robustness and accuracy. A fully self-supervised learning approach for estimating depth and ego-motion from monocular thermal video is proposed in \cite{ShinMaximizing}, introducing a temporally consistent mapping technique to enhance contrast and structural information. A semi-direct VO system that fuses raw thermal and depth images with a dedicated NUC handling module for sensor disruption recovery is proposed in \cite{10111061}. An end-to-end navigation pipeline using LWIR imagery and the deep learning model TrajNet for trajectory prediction under model predictive control is proposed in \cite{10611311}, enabling reliable nighttime operation. Finally, a monocular thermal visual odometry method for outdoor environments is proposed in \cite{7399731}, addressing scale ambiguity through road segmentation and mitigating NUC-induced pose estimation failures via a predictive trigger strategy.

The mentioned works highlight diverse strategies for addressing key challenges in thermal imaging, including low texture, NUC interruptions, and dynamic object interference, while extending navigation capabilities to low-light and GPS-denied environments. Our contribution diverges from these prior works in two ways. First, our approach simultaneously generates dense depth maps and enhances thermal images, enabling metric scale recovery and allowing for direct integration into existing SLAM frameworks without modification. Second, we present a non-radiometric thermal–depth \ac{UAV} dataset, demonstrating robustness in fully dark indoor environments where conventional RGB-based methods fail. By incorporating recurrent modeling for temporal consistency and ensuring real-time deployment on embedded hardware, our framework emphasizes both the practical applicability and generalizability of thermal-based navigation.

\section{Methodology}\label{sec:methodology}



Non-radiometric thermal imagery presents unique challenges for depth estimation and SLAM, including low contrast, high dynamic range, and weak structural cues that hinder reliable feature extraction. To overcome these limitations, we propose a lightweight preprocessing network, T-RefNet, that refines thermal inputs and enhances their structural visibility. Integrated with a \ac{RB} and a supervised depth decoder, the proposed pipeline enables temporally consistent and geometrically accurate depth predictions from thermal-only sequences, thereby facilitating robust SLAM performance. As illustrated in Fig.~\ref{fig:method_pipeline}, the proposed system takes as input a raw 16-bit thermal image captured by an LWIR camera. To compensate for the inherently low contrast and high dynamic range of thermal data, a lightweight convolutional module, T-RefNet, is introduced to refine and normalize the input data. This module produces two complementary outputs: i) a normalized thermal image that serves as input to the supervised depth estimation backbone, and ii) an 8-bit color-mapped representation suitable for reliable ORB feature extraction within ORB-SLAM3. By providing both depth priors and texture-rich images, the system overcomes the limitations of raw thermal imagery, enabling robust SLAM operation with metric scale recovery.

\subsection{Radiometric and non-radiometric thermal camera}
\label{subsec:Radiometric}

Thermal imaging systems are either radiometric, delivering calibrated per-pixel temperatures for quantitative analysis, or non-radiometric, providing only relative contrast. Non-radiometric outputs are auto-scaled by frame content and internal temperature, so pixel values lack consistent physical meaning \cite{flir_radiometric}. On the other hand, non-radiometric thermal cameras are low-cost, more accessible, and do not require continuous thermal calibration, while still providing sufficient relative contrast for navigation-focused tasks.

\begin{figure}[t]
  \centering
  \setlength{\tabcolsep}{0pt}
  \renewcommand{\arraystretch}{1.0}

  \begin{tabular}{@{}c@{\hspace{0.03\linewidth}}c@{}}
    \parbox[c]{0.49\linewidth}{
      \centering
      \subfloat[Radiometric and non-radiometric\label{fig:thermal_a}]{%
        \includegraphics[width=0.98\linewidth]{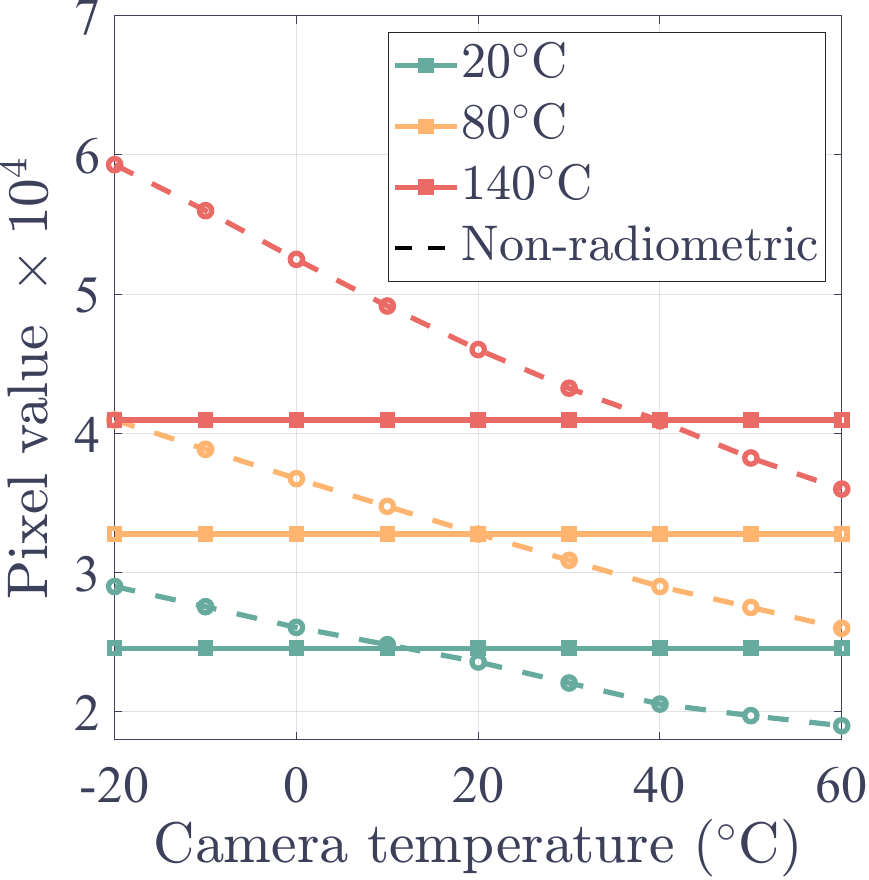}}%

    }
    &
    \parbox[c]{0.48\linewidth}{
      \centering
      \subfloat[Thermal enhancement techniques\label{fig:thermal_b}]{%
        \begin{tabular}{@{}c@{\hspace{0.01\linewidth}}c@{}}
          \includegraphics[width=0.49\linewidth]{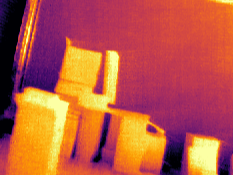} &
          \includegraphics[width=0.49\linewidth]{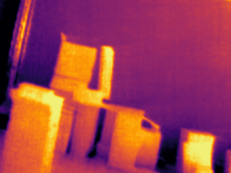} \\
          {\scriptsize i-) Raw thermal} & {\scriptsize ii-) Smoothed} \\[3pt]
          \includegraphics[width=0.49\linewidth]{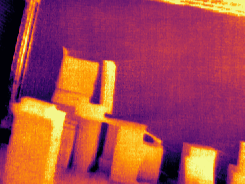} &
          \includegraphics[width=0.49\linewidth]{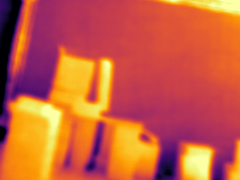} \\
          {\scriptsize iii-) \ac{CLAHE}} & {\scriptsize iv-) T-RefNet} \\
        \end{tabular}
      }
    }
  \end{tabular}

  \caption{Comparison of the thermal image preprocessing methods. 
  (a) Radiometric vs. non-radiometric thermal cameras at different \ac{TBB} values. 
  Solid lines represent radiometric outputs, while dashed lines indicate non-radiometric behavior.\protect\footnotemark[1] 
  (b) Thermal image enhancement techniques: 
  i) Raw input suffers from noise that disrupts gradients; 
  ii) Gaussian smoothing reduces noise but blurs edges; 
  iii) \ac{CLAHE} boosts local contrast but introduces spurious keypoints; 
  iv) T-RefNet preserves edges while denoising, yielding stable features for SLAM.}
  \label{fig:thermal_combined}
\end{figure}

Figure~\ref{fig:thermal_a} compares pixel responses of radiometric and non-radiometric cameras across varying \ac{TBB} temperatures. Radiometric thermal cameras produce consistent, near-linear outputs
, enabling temperature-aware preprocessing such as \ac{CLAHE} or adaptive thresholds. By contrast, non-radiometric cameras re-map intensities frame by frame, causing histogram shifts, abrupt jumps with hot/cold regions, and unstable normalization.

To enhance thermal imagery for downstream vision tasks, different methods are compared in Fig.~\ref{fig:thermal_b}. Raw 8-bit frames contain high-frequency noise, Gaussian smoothing reduces noise but blurs salient edges, and \ac{CLAHE} improves contrast while amplifying spurious features. In contrast, the CNN-based T-RefNet produces denoised yet structurally consistent outputs, preserving contours and enabling stable feature extraction for SLAM.

\footnotetext[1]{FLIR Boson: \url{https://oem.flir.com/products/boson}}

\subsection{Training flow of the depth estimation}

\begin{algorithm}[!h]
\label{alg:train}
\caption{Training flow of the refinement--sequence depth estimation.}
\KwIn{Thermal sequences
$\{x_t\}_{t=1}^T$, and the ground-truth depths $\{z^{gt}_t\}_{t=1}^T$\ 
\\1. Initialize parameters:
$\theta$ (T-RefNet), $\phi$ (Enc--Dec),
$\psi$ (RB), lr $\eta$}

\For{$t \gets 1$ \KwTo $T$}{
    $y_t \gets f_{\theta}(x_t)$\;
    $\{h^{enc}_{t,\ell}\}_{\ell=0}^{L} \gets \text{Encoder}_{\phi}(y_t)$\;
    $h^{in}_t \gets \mathrm{F_{latent.S}}(h^{enc}_{t,L})$\;
    $h^{RB}_t \gets \mathrm{W}_{\psi}(h^{in}_t, h^{RB}_{t-1})$\;
    $h^{enc}_{t,L} \gets \mathrm{F_{readout}}(h^{RB}_t)$\;
    $\hat{z}_t \gets \text{Decoder}_{\phi}\!\big(\{h^{enc}_{t,\ell}\}_{\ell=0}^{L}\big)$\;
}
2. Calculate loss function\\
$\mathcal{L}_{\text{total}} =
\lambda_1 \mathcal{L}_{\text{SIlog}}
+ \lambda_2 \mathcal{L}_{\text{SSIM}}
+ \lambda_3 \mathcal{L}_{\text{ord}}
+ \lambda_4 \mathcal{L}_{\text{sm}}
$\;
3. Update the weights:\\
$\theta \gets \theta - \eta \nabla_{\theta}\mathcal{L}_{\text{total}}$\;

$\phi   \gets \phi   - \eta \nabla_{\phi}\mathcal{L}_{\text{total}}$\;
$\psi   \gets \psi   - \eta \nabla_{\psi}\mathcal{L}_{\text{total}}$\;

\end{algorithm}

\noindent
The training procedure for the proposed T-RefNet-based sequence depth estimation model is described in Algorithm~\ref{alg:train}. At each timestep, the input thermal frame is first refined by the T-RefNet module and then encoded into multi-scale features. These features are passed through the \ac{RB} to capture temporal context and finally decoded into a depth map. 

The model parameters are updated end-to-end with a composite loss designed to enforce both geometric consistency and perceptual accuracy. Building on the concept of combined loss formulations from prior work~\cite{ShinMaximizing}, we extend this idea by integrating multiple complementary objectives into a single framework. Specifically, the loss includes:
i) a scale-invariant term $\mathcal{L}_{\text{SIlog}}$~\cite{eigen2014depth}, assigned the largest weight ($0.9$) to capture the global depth structure;  
ii) a perceptual similarity term $\mathcal{L}_{\text{SSIM}}$~\cite{godard2019digging}, weighted $0.4$ to preserve local structures and textures;  
iii) a depth-ordering term $\mathcal{L}_{\text{ord}}$~\cite{Xian2020StructureRanking}, weighted $0.1$ to enforce correct relative ordering between pixels; and  
iv) an edge-aware smoothness term $\mathcal{L}_{\text{sm}}$~\cite{edge_aware}, also weighted $0.1$, to regularize depth predictions while respecting image boundaries. This formulation improves stability and accuracy in thermal depth estimation.


To maintain efficiency and real-time capability, the encoder backbone is instantiated with lightweight architectures such as EfficientNet-B0, MobileNet, or ResNet-8, offering a favorable trade-off between accuracy and computational cost. 
For temporal modeling, the refined thermal sequence is further processed by either a ConvGRU bottleneck or a reservoir computing based network, 
both integrated into the depth estimation network. These recurrent architectures capture frame-to-frame dependencies and enforce temporal consistency across predictions, which is essential for stable SLAM operation in dynamic or texture-poor thermal environments. As a result, SLAM initialization becomes more reliable, mapping accuracy improves, and real-time tracking performance is enhanced.

\subsection{Reservoir computing}
RC constitutes a recurrent neural network paradigm that represents and processes temporal and sequential data~\cite{jaeger2001echo}.
Fundamentally, RC operates by embedding input signals into a high-dimensional state space via a randomly connected recurrent network of non-linear neurons. This state space inherently captures the temporal dependencies of the input, while the readout layer maps the reservoir dynamics onto the desired output.

Let $\mathbf{u}(t) \in \mathcal{R}^K$ be the input at time $t$ with $K$ input neurons. The internal state of the reservoir, $\mathbf{x}(t)\in \mathcal{R}^N$, is expressed as 
\begin{equation}
    \mathbf{x}(t+1) = f \left(\mathbf{W}_{in}\mathbf{u}(t+1) + \mathbf{W}\mathbf{x}(t)\right) ,
\end{equation}
where $f$ is  a sigmoidal function, $\mathbf{W}_{in}\in \mathcal{R}^{N\times K}$ represents the input matrix, $\mathbf{W}\in \mathcal{R}^{N\times N}$ the weight matrix of the reservoir, and $\mathbf{y}\in \mathcal{R}^L$ is the output signal. 
Then the output is computed as 
\begin{equation}
     \mathbf{y}(t+1) = f^{out}\left(\mathbf{W}_{out} \mathbf{x}(t+1)\right) ,
\end{equation}
with $\mathbf{W}_{out} \in \mathcal{R}^{L\times N}$. We base our reservoir implementation on ~\cite{hasani2021liquid}, which uses a biologically realistic representation of neurons, namely the \ac{lif} neuron. The reservoir layer consists of a vector of membrane potentials of $N$ of excitatory and inhibitory \ac{lif} neurons $\mathbf{v}(t) \in \mathcal{R}^N$. 
A differential equation describes the \ac{lif} neuron as~\cite{gerstner2014neuronal}:
\begin{equation}
    \tau_m \frac{d V(t)}{dt} = - V(t) + R_m I(t),
\end{equation}
where  $V(t)$ is the membrane potential at time $t$, $\tau_m$ is the membrane time constant $R_m$ is the membrane resistance and $I(t)$ is the input current at time $t$.

\section{Experiments}\label{sec:experiment}

In this section, we present the thermal-to-depth estimation results of the proposed model, followed by an evaluation of its integration into ORB-SLAM3 across various trajectories and scenes, using both \ac{UAV} and handheld devices to highlight the advantages for robust localization.

\subsection{Evaluation and baselines}
To comprehensively evaluate the proposed approach, we conducted experiments on two different thermal–depth datasets: (i) the indoor-dark subset of VIVID++\cite{vivid_dataset}, which is recorded with a radiometric thermal camera, and (ii) a custom dataset collected with a non-radiometric thermal sensor and a depth camera. This dual evaluation setup enables us to assess performance under both radiometric and non-radiometric conditions. The thermal data were captured with a Flir Boson+\protect\footnotemark[1] non-radiometric shuttered camera (640$\times$512).  The dataset comprises approximately 65,000 samples, covering diverse lighting conditions—bright, dark, and semi-lit—and including scenes with both hot and cold objects to improve robustness across thermal distributions.

For comparison, we include both RGB-trained depth estimation networks (ZoeDepth~\cite{zoedepth}, DepthAnything-V2~\cite{yang2024depthv2}) and thermal-specific approaches from the literature \cite{9662239,ShinMaximizing,Kong_2025,10890218}. Since ZoeDepth and DepthAnything-V2 were trained on RGB images, we pre-processed our thermal inputs by mapping them to RGB format before inference. Regarding \cite{ShinMaximizing}, we retrained and evaluated the model on our non-radiometric dataset using the sequences we collected.
We use key metrics to quantitatively evaluate our  methods against the baselines, such as mean absolute relative error (AbsRel), root mean squared error (RMSE), and accuracy under thresholds $1.25, 1.25^2, 1.25^3$ (a1, a2, a3).

\subsection{Thermal-to-depth estimation results}

\begin{figure*}[!b]
\centering
\setlength{\tabcolsep}{0pt}             
\renewcommand{\arraystretch}{0.98}       
\newcommand{\colw}{0.158\textwidth}      
\newcommand{\rowlab}[2]{\raisebox{#1}{\rotatebox[origin=c]{90}{\textbf{\footnotesize #2}}}}

\begin{tabular}{@{}m{0.03\textwidth}*{6}{m{\colw}}@{}}

\rowlab{-3em}{VIVID++} &  
\begin{minipage}[t]{\colw}\centering
  \includegraphics[width=\linewidth]{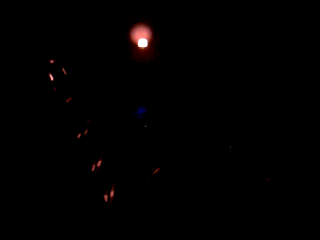}\\[-2pt]
  \includegraphics[width=\linewidth]{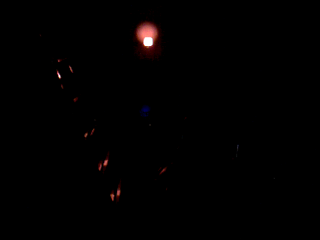}
\end{minipage} &
\begin{minipage}[t]{\colw}\centering
  \includegraphics[width=\linewidth]{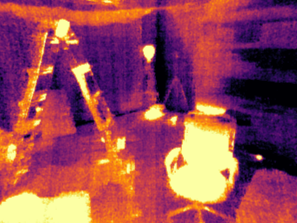}\\[-2pt]
  \includegraphics[width=\linewidth]{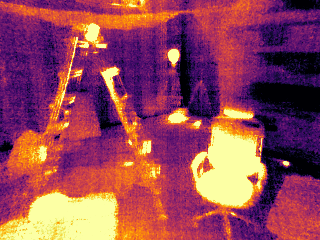}
\end{minipage} &
\begin{minipage}[t]{\colw}\centering
  \includegraphics[width=\linewidth]{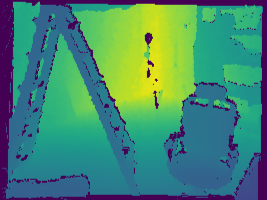}\\[-2pt]
  \includegraphics[width=\linewidth]{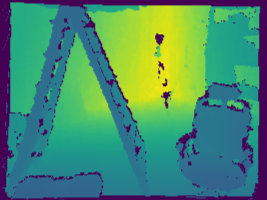}
\end{minipage} &
\begin{minipage}[t]{\colw}\centering
  \includegraphics[width=\linewidth]{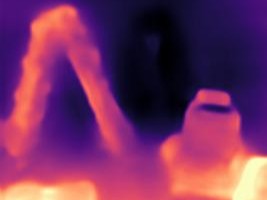}\\[-2pt]
  \includegraphics[width=\linewidth]{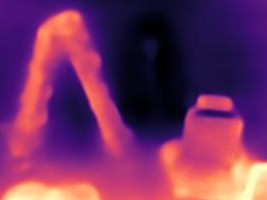}
\end{minipage} &
\begin{minipage}[t]{\colw}\centering
  \includegraphics[width=\linewidth]{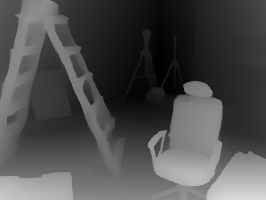}\\[-2pt]
  \includegraphics[width=\linewidth]{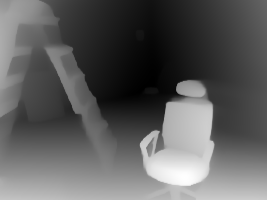}
\end{minipage} &
\begin{minipage}[t]{\colw}\centering
  \includegraphics[width=\linewidth]{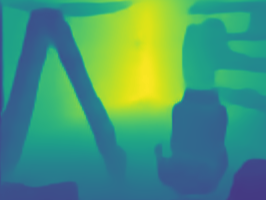}\\[-2pt]
  \includegraphics[width=\linewidth]{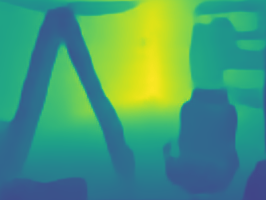}
\end{minipage}
\\[-2ex] 

\rowlab{-3em}{Our Dataset} &
\subfloat[RGB\label{fig:rgb}]{%
  \begin{minipage}[t]{\colw}\centering
    \includegraphics[width=\linewidth]{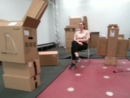}\\[-2pt]
    \includegraphics[width=\linewidth]{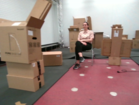}
  \end{minipage}} &
\subfloat[Thermal\label{fig:thermal}]{%
  \begin{minipage}[t]{\colw}\centering
    \includegraphics[width=\linewidth]{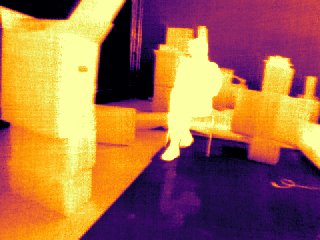}\\[-2pt]
    \includegraphics[width=\linewidth]{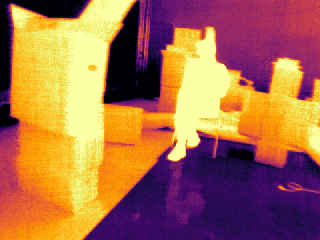}
  \end{minipage}} &
\subfloat[Ground Truth\label{fig:gt}]{%
  \begin{minipage}[t]{\colw}\centering
    \includegraphics[width=\linewidth]{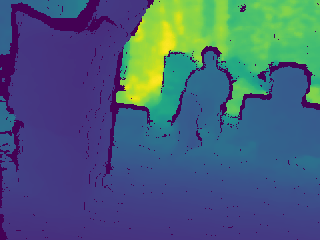}\\[-2pt]
    \includegraphics[width=\linewidth]{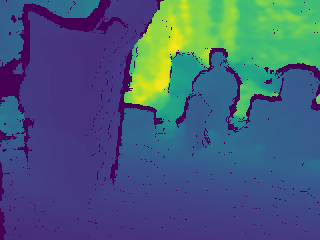}
  \end{minipage}} &
\subfloat[Shin et al.~\cite{ShinMaximizing} \label{fig:shin}]{%
  \begin{minipage}[t]{\colw}\centering
    \includegraphics[width=\linewidth]{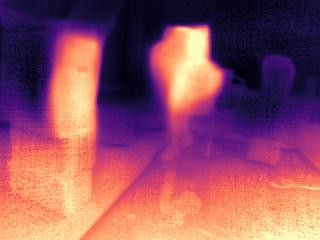}\\[-2pt]
    \includegraphics[width=\linewidth]{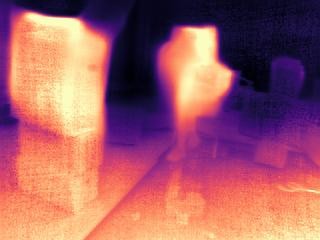}
  \end{minipage}} &
\subfloat[DepthAnything~\cite{yang2024depthv2}\label{fig:da2}]{%
  \begin{minipage}[t]{\colw}\centering
    \includegraphics[width=\linewidth]{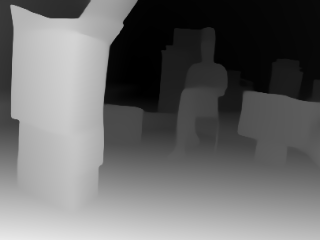}\\[-2pt]
    \includegraphics[width=\linewidth]{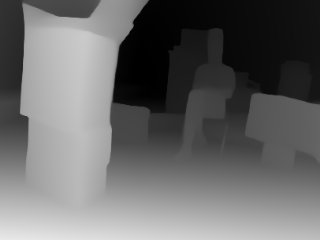}
  \end{minipage}} &
\subfloat[Ours (RC+Eff-B0)\label{fig:pred}]{%
  \begin{minipage}[t]{\colw}\centering
    \includegraphics[width=\linewidth]{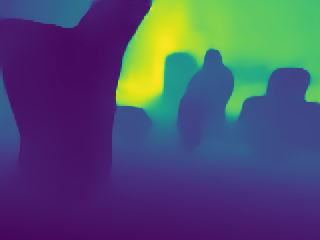}\\[-2pt]
    \includegraphics[width=\linewidth]{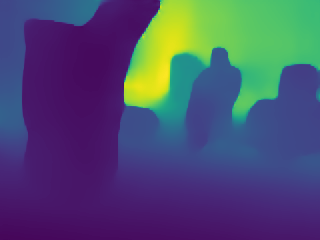}
  \end{minipage}}
\end{tabular}

\caption{Qualitative comparison across two datasets. Top: VIVID++; bottom: our dataset. Each row shows two temporally adjacent frames. Columns: (a) RGB, (b) thermal, (c) thermal-aligned ground-truth depth, (d) Shin et al.~\cite{ShinMaximizing}, (e) DepthAnything-V2~\cite{yang2024depthv2} (RGB-only), (f) Our representative proposed model with \ac{RC}.}
\label{fig:comparison}
\end{figure*}

\begin{table}[!t]
\centering
\caption[Depth estimation comparison on indoor-dark (VIVID)]{Quantitative comparison of depth estimation accuracy across different architectures on the indoor-dark subset of the VIVID++\cite{vivid_dataset} dataset. Best values are shown in bold.}
\begin{threeparttable}
\resizebox{\columnwidth}{!}{%
\begin{tabular}{lccccc}
\toprule
\textbf{Model} & \textbf{AbsRel} & \textbf{RMSE} & \textbf{a1} & \textbf{a2} & \textbf{a3} \\
\midrule
Shin (T)~\cite{9662239}       & 0.232 & 0.740 & 0.618 & 0.907 & 0.987 \\
Shin (MS)~\cite{9662239}      & 0.166 & 0.566 & 0.768 & 0.967 & 0.994 \\
Shin (Max.)~\cite{ShinMaximizing} & 0.149 & 0.517 & 0.813 & 0.969 & 0.994 \\
ZoeDepth~\cite{zoedepth}         & 0.165 & 0.533 & 0.788 & 0.944 & 0.991 \\
DepthAnything-V2~\cite{yang2024depthv2} & 0.112 & 0.378 & 0.902 & 0.970 & 0.990 \\
Ye et al.~\cite{10890218}      & 0.145 & 0.499 & 0.827 & 0.969 & 0.994 \\
MSDFNet~\cite{Kong_2025}       & 0.139 & 0.470 & 0.847 & \textbf{0.980} & \textbf{0.996} \\
\midrule
Eff-B0 noRB (ours)        & 0.139  & 0.497  & 0.839 & 0.945 & 0.984 \\
Eff-B0+GRU noTRN (ours)   & 0.079  & 0.325  & 0.929 & \textbf{0.980} & 0.995 \\
ResNet8+GRU (ours)    & 0.079  & 0.345  & 0.913 & 0.970 & 0.990 \\
MobileNet+GRU (ours)   & 0.072  & 0.318  & 0.928 & 0.977 & 0.993 \\
Eff-B0+GRU (ours)         & \textbf{0.063} & \textbf{0.298} & \textbf{0.940} & \textbf{0.980} & 0.993 \\
Eff-B0+RC (ours)          & 0.069  & 0.313  & 0.931 & 0.976 & 0.993 \\
\bottomrule
\end{tabular} }
\begin{tablenotes}
\footnotesize
\item Eff-B0: EfficientNet-B0 backbone; 
noRB: without recurrent block; 
\item GRU: ConvGRU; 
noTRN: without T-RefNet.
\end{tablenotes}
\end{threeparttable}
\label{tab:dark_vivid_results}
\end{table}

Table~\ref{tab:dark_vivid_results} shows the quantitative results on the VIVID++ indoor-dark subset. The methods demonstrate competitive performance, with MSDFNet~\cite{Kong_2025} achieving the best $a2$ (0.980) and $a3$ accuracy (0.996). Among general-purpose RGB models, DepthAnything-V2~\cite{yang2024depthv2} yields strong results ($\text{AbsRel}=0.112$, $\text{RMSE}=0.378$). However, as illustrated in Fig.~\ref{fig:da2}, its predictions are not entirely consistent: Although it preserves sharpness in some static scenes, during motion it often blurs structures, removes objects, and hinders accurate depth analysis. In contrast, our  architectures significantly outperform all baselines on most metrics. In particular, our model with the EfficientNet-B0 encoder achieves the best results with $\text{AbsRel}=0.063$, $\text{RMSE}=0.298$, and $a1=0.940$, highlighting the effectiveness of the combined ConvGRU and T-RefNet modules. In addition, the \ac{RC} variant delivers results close to the best model in the radiometric data set, while using only about 50k parameters (including its latency space block and readout) with 32 reservoir neurons  compared to over 800k parameters required for ConvGRU and its corresponding components, making it a lightweight yet competitive alternative.

\begin{table}[!t]
\centering
\caption[Our dataset Depth estimation]{Evaluation results of thermal-to-depth networks on a custom indoor dataset acquired with nonradiometric thermal and depth sensors. Best values are shown in bold.}
\resizebox{\columnwidth}{!}{%
\begin{tabular}{lccccc}
\toprule
\textbf{Model} & \textbf{AbsRel} & \textbf{RMSE} & \textbf{a1} & \textbf{a2} & \textbf{a3} \\
\midrule
Shin (Max.)~\cite{ShinMaximizing}   & 0.262 & 1.273 & 0.589 & 0.890 & 0.960 \\
ZoeDepth~\cite{zoedepth}            & 0.243 & 1.110 & 0.605 & 0.885 & 0.954 \\
DepthAnything-V2~\cite{yang2024depthv2} & 0.267 & 1.043 & 0.571 & 0.863 & 0.931 \\
\midrule
ResNet8+GRU (ours)     & 0.109 & 0.516  & 0.886 & 0.943 & 0.969 \\
MobileNet+GRU (ours)   & 0.085  & 0.453  & 0.911 & 0.951 & 0.971 \\
Eff-B0+GRU (ours)     & 0.079 & \textbf{0.424}  & 0.920 & 0.955 & 0.971 \\
Eff-B0+RC (ours)     & \textbf{0.076} & 0.439  & \textbf{0.929} & \textbf{0.965} & \textbf{0.981} \\
\bottomrule
\end{tabular} }
\label{tab:rat_results}
\end{table}

Table~\ref{tab:rat_results} presents the evaluation on the proposed non-radiometric dataset, which is considerably more challenging due to fluctuating pixel intensities caused by auto-scaling and internal heating. The models trained purely on RGB inputs (ZoeDepth, DepthAnything-V2) perform poorly, with higher $\text{AbsRel}$ and lower accuracies. Regarding \cite{ShinMaximizing}, as illustrated in Fig.~\ref{fig:shin}, the method employs radiometric-specific preprocessing and performs reasonably on the VIVID++ dataset, but on non-radiometric data its consistency degrades when hot or cold objects enter or leave the scene. In contrast, our  models remain robust, with \ac{RC} combined with EfficientNet-B0 giving slightly better results on the non-radiometric dataset ($\text{AbsRel}=0.076$, $a1=0.929$), as also shown in Fig.~\ref{fig:comparison}. Although the difference compared to the GRU-based model is not large, it is notable that the \ac{RC} variant, with a lighter architecture, performs better in the more variable non-radiometric conditions. These findings underscore the importance of radiometric invariance and demonstrate that our approach generalizes well across both radiometric and non-radiometric settings.

\subsection{Localization results}
\begin{figure}[!b]
\centering
\setlength{\tabcolsep}{2pt}
\renewcommand{\arraystretch}{0.0}
\begin{tabular}{c|c|c|c} 
    & t = 35.0 s & t = 35.3 s & t = 35.6 s
    \\[+0.15em]
    \hline \\[+0.1em]
    \rotatebox{90}{\hspace{1.9em}RGB} &
    \includegraphics[width=0.14\textwidth]{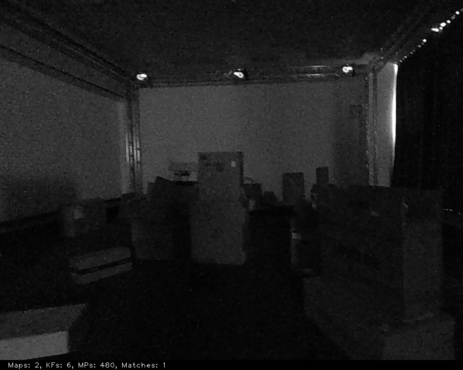} &
    \includegraphics[width=0.14\textwidth]{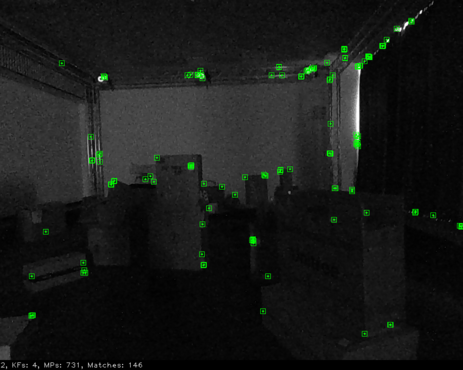} &
    \includegraphics[width=0.14\textwidth]{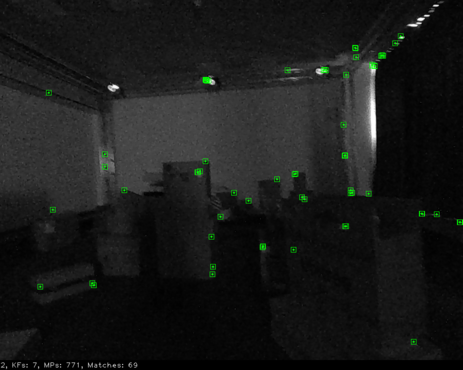} 
    \\[+0.1em]
    \hline \\[+0.1em]
    \rotatebox{90}{\hspace{1.0em}Thermal} &
    \includegraphics[width=0.14\textwidth]{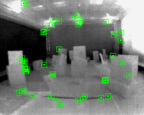} &
    \includegraphics[width=0.14\textwidth]{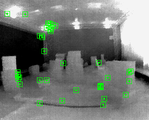} &
    \includegraphics[width=0.14\textwidth]{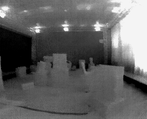} \\[+0.1em]
    \hline \\[+0.1em]
    \rotatebox{90}{\hspace{1.0em}T-RefNet} &
    \includegraphics[width=0.14\textwidth]{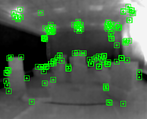} &
    \includegraphics[width=0.14\textwidth]{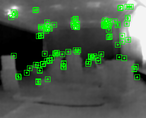} &
    \includegraphics[width=0.14\textwidth]{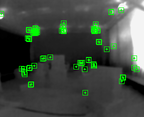} \\
\end{tabular}
\caption{Feature tracking results of ORB-SLAM3 using different image inputs: RGB images (top row), raw 8-bit thermal images (middle row), and T-RefNet enhanced thermal images (bottom row). 
While RGB features degrade under low-light indoor conditions, raw thermal inputs suffer from noise and low contrast. 
In contrast, T-RefNet outputs provide more stable and repeatable features, leading to improved tracking robustness.}
\label{fig:thermal_rgb_comparison}
\end{figure}

To evaluate the practical applicability of our thermal-based preprocessing and depth estimation pipeline within a visual SLAM framework, we integrated the outputs of both the T-RefNet and the depth prediction network into the ORB-SLAM3 framework. The goal is to evaluate how well these outputs support localization and mapping under thermal-only input conditions.

All experiments are conducted offline using ROS bag files. Thermal and RGB-D images were captured and stored in real time, and the synchronized data streams were recorded into `.bag` files. The stored bags were then played back for evaluation to ensure consistent and reproducible conditions.

In low-light indoor scenarios, the quality of input images directly affects the ability of ORB-SLAM3 to maintain reliable tracking. As illustrated in Fig.~\ref{fig:thermal_rgb_comparison}, directly converting raw 16-bit thermal images into an 8-bit format results in frequent loss of structural features due to low contrast and high noise levels. The detected features are too sparse and unstable to support consistent tracking, making it impossible for ORB-SLAM3 to generate a meaningful trajectory. In dark illumination conditions, RGB images also fail to provide sufficient structure for reliable feature extraction, as shown in the top row of Fig.~\ref{fig:thermal_rgb_comparison}. In contrast, T-RefNet thermal frames maintain edge consistency and enhance feature visibility, facilitating stable keypoint detection and accurate pose estimation. Consequently, except for evaluations in bright environments, only the T-RefNet image was used in dark scenarios, while raw 8-bit thermal and RGB inputs were excluded from quantitative analysis.

\begin{figure}[!b]
  \centering
\setlength{\tabcolsep}{2pt}
\renewcommand{\arraystretch}{1}

\begin{tabular}{m{0.018\linewidth} c c c}

&
\subfloat[Test setup]{\includegraphics[width=0.305\linewidth]{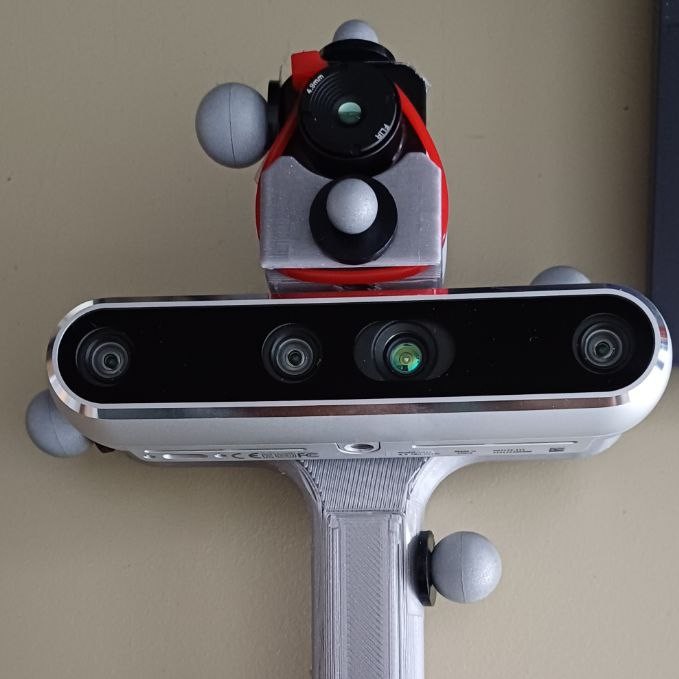}\label{fig:line:a}}
&
\subfloat[Scene]{\includegraphics[width=0.305\linewidth]{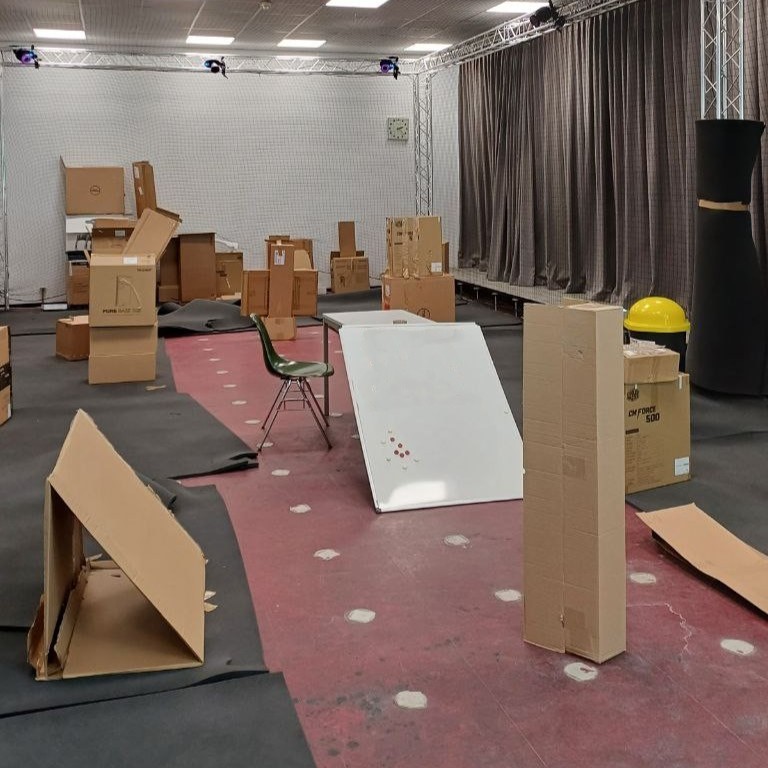}\label{fig:line:bright_line}}
&
\subfloat[Trajectory]{\includegraphics[width=0.305\linewidth]{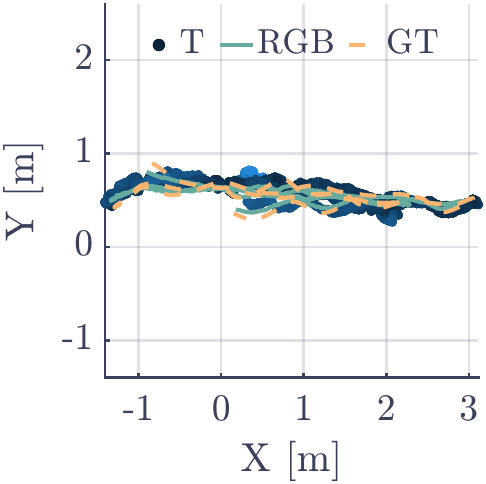}\label{fig:line:c}}
\\[-0.4ex]

\centering
\raisebox{8mm}[0pt][0pt]{%
\rotatebox{90}{%
{\color[HTML]{3D405B}\fontsize{7pt}{7pt}\selectfont Position [m]}}}
&
\subfloat[X position]{\includegraphics[width=0.295\linewidth]{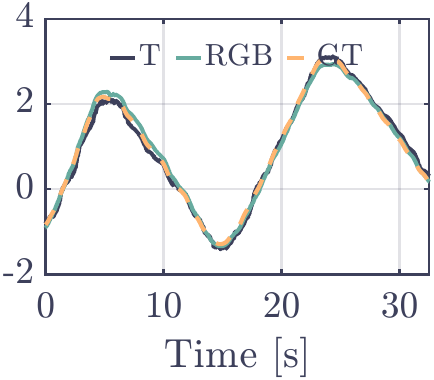}\label{fig:line:d}}
&
\subfloat[Y position]{\includegraphics[width=0.305\linewidth]{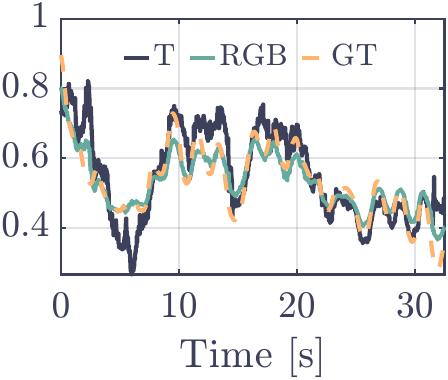}\label{fig:line:e}}
&
\subfloat[Z position]{\includegraphics[width=0.305\linewidth]{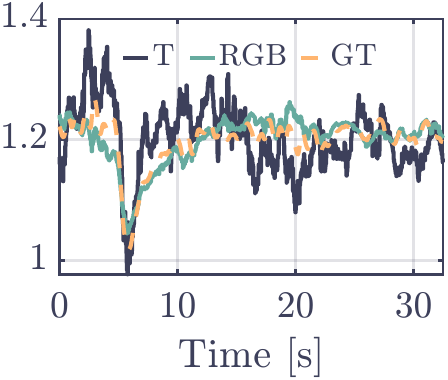}\label{fig:line:f}}
\\[-0.4ex]

\centering
\raisebox{7mm}[0pt][0pt]{%
\rotatebox{90}{%
{\color[HTML]{3D405B}\fontsize{6.5pt}{7pt}\selectfont Abs. Error [m]}}}
&
\subfloat[X position]{\includegraphics[width=0.305\linewidth]{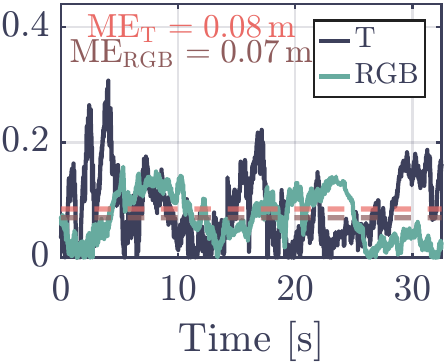}\label{fig:line:g}}
&
\subfloat[Y position]{\includegraphics[width=0.305\linewidth]{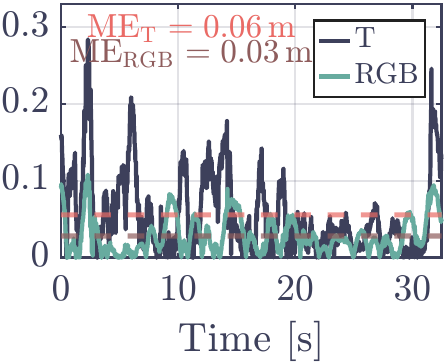}\label{fig:line:h}}
&
\subfloat[Z position]{\includegraphics[width=0.305\linewidth]{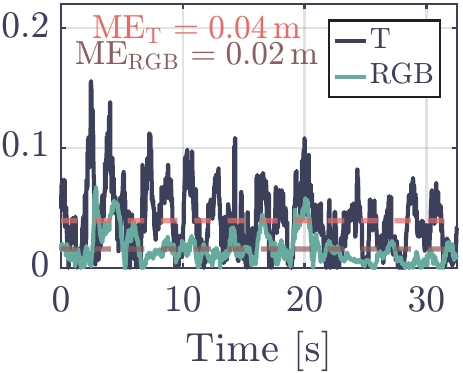}\label{fig:line:i}}

\end{tabular}

  \caption[ORB-SLAM3 trajectory and error analysis in bright scene with handheld device]{Experimental results in a bright indoor scene with a handheld device. (a) Test setup with RGB-D and thermal cameras, (b) sample scene view, and (c) ground-truth trajectory. (d–f) Estimated trajectories along the X, Y, and Z axes from ORB-SLAM3 with RGB-D and T-RefNet refined thermal input are compared against ground truth. (g–i) Absolute position errors along each axis are shown, with mean error (ME) values highlighted for both RGB-D and T-RefNet inputs.}
  \label{fig:line}
\end{figure}

Three evaluation scenarios are designed to assess the proposed pipeline under varying conditions: i) a bright environment with abundant features and linear motion, comparing RGB-D and estimated thermal depth; ii) a dark environment with circular motion and fewer features, evaluating thermal depth performance; and iii) a \ac{UAV}-based test across corridors with varying illumination, where each corridor presents different lighting conditions.

In the bright environment scenario with linear back-and-forth motion (Fig.~\ref{fig:line}), RGB-D naturally achieves higher accuracy than thermal depth due to the abundance of detectable features. Nevertheless, since the motion is simple and the scene provides rich structural cues, both methods produce stable trajectories. The handheld setup further reduces motion jitter, resulting in consistent tracking across all axes. Quantitatively, the Euclidean mean error is about 0.11~m for thermal depth and about 0.08~m for RGB-D indicating that while RGB-D has an advantage, the T-RefNet–enhanced thermal input with estimated depth still provides sufficiently accurate estimates for reliable localization.

\begin{figure}[!t]
  \centering
\setlength{\tabcolsep}{2pt}
\renewcommand{\arraystretch}{1}
\begin{tabular}{m{0.02\linewidth} c c c}

&
\subfloat[Test setup]{\includegraphics[width=0.3\linewidth]{figures/result/wand.jpg}\label{fig:circle:a}}
&
\subfloat[Scene]{\includegraphics[width=0.3\linewidth]{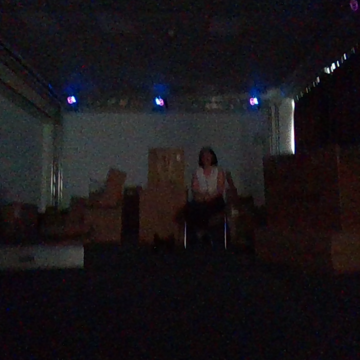}\label{fig:circle:dark_scene}}
&
\subfloat[Trajectory]{\includegraphics[width=0.30\linewidth]{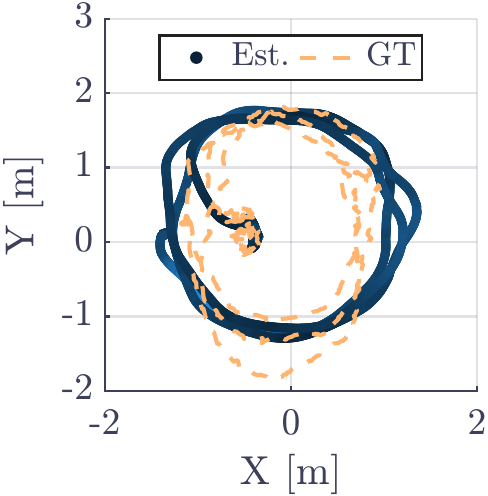}\label{fig:circle:c}}
\\[-0.3ex]

\centering
\raisebox{8mm}[0pt][0pt]{%
\rotatebox{90}{%
{\color[HTML]{3D405B}\fontsize{7pt}{7pt}\selectfont Position [m]}}}
&
\subfloat[X position]{\includegraphics[width=0.305\linewidth]{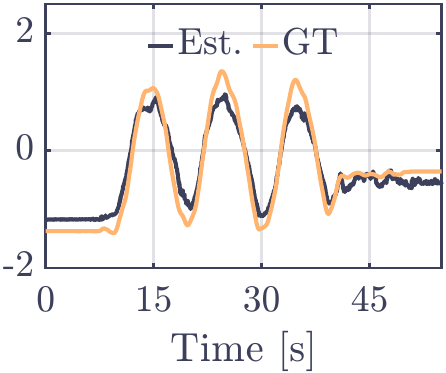}\label{fig:circle:d}}
&
\subfloat[Y position]{\includegraphics[width=0.305\linewidth]{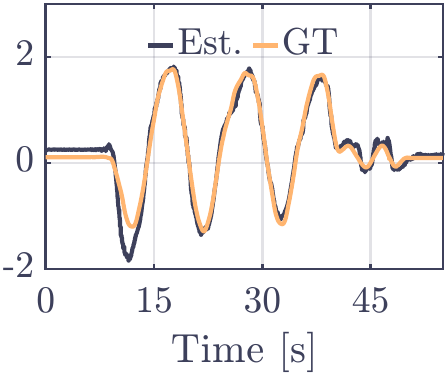}\label{fig:circle:e}}
&
\subfloat[Z position]{\includegraphics[width=0.295\linewidth]{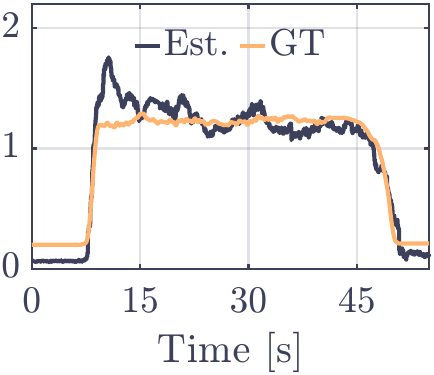}\label{fig:circle:f}}
\\[-0.3ex]

\centering
\raisebox{6mm}[0pt][0pt]{%
\rotatebox{90}{%
{\color[HTML]{3D405B}\fontsize{6.5pt}{7pt}\selectfont Abs. Error [m]}}}
&
\subfloat[X abs. error]{\includegraphics[width=0.305\linewidth]{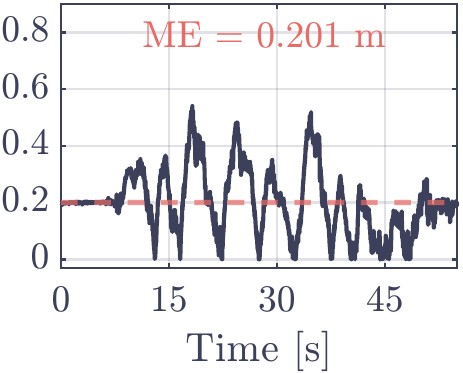}\label{fig:circle:g}}
&
\subfloat[Y abs. error]{\includegraphics[width=0.305\linewidth]{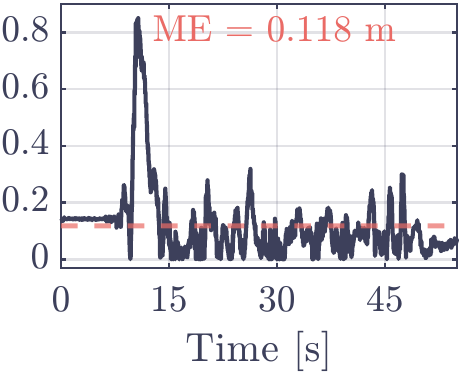}\label{fig:circle:h}}
&
\subfloat[Z abs. error]{\includegraphics[width=0.305\linewidth]{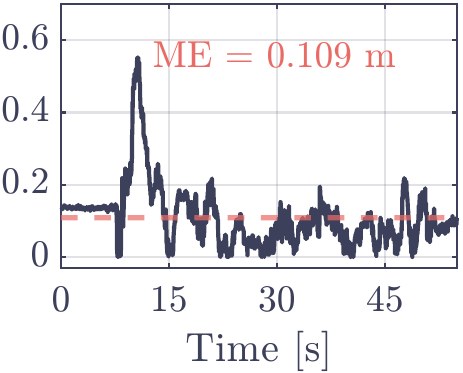}\label{fig:circle:i}}

\end{tabular}

  \caption[ORB-SLAM3 trajectory and error analysis in dark scene with handheld device]{Experimental results in a dark indoor scene with a handheld device performing circular motion. (a) Handheld device, (b) sample scene view, and (c) ground-truth trajectory. (d–f) Estimated trajectories along the X, Y, and Z axes from ORB-SLAM3 with T-RefNet are compared against ground truth. (g–i) Absolute position errors are shown for each axis, with ME values highlighted.}
  \label{fig:circle}
\end{figure}

\begin{figure}[t]
  \centering

\setlength{\tabcolsep}{2pt}
\renewcommand{\arraystretch}{1}

\begin{tabular}{m{0.015\linewidth} c c c}

&
\subfloat[Test setup]{\includegraphics[width=0.30\linewidth]{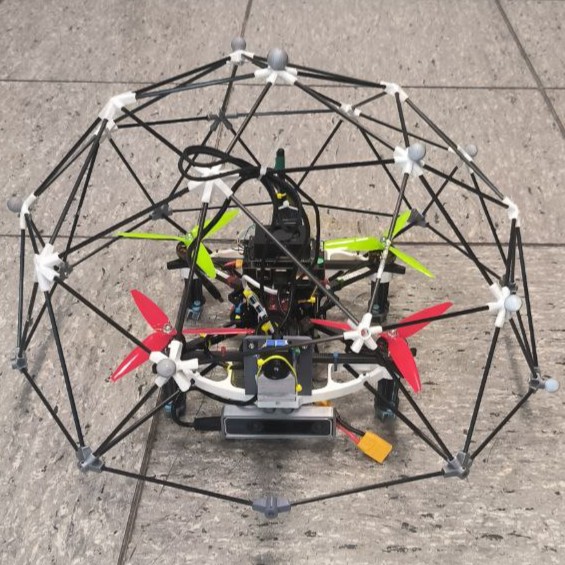}\label{fig:drone-grid:a}}
&
\subfloat[Scene]{\includegraphics[width=0.30\linewidth]{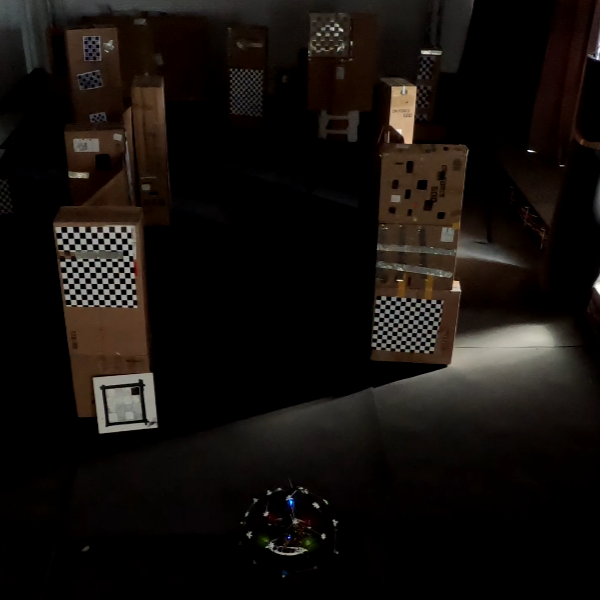}\label{fig:drone-grid:b}}
&
\subfloat[Trajectory]{\includegraphics[width=0.30\linewidth]{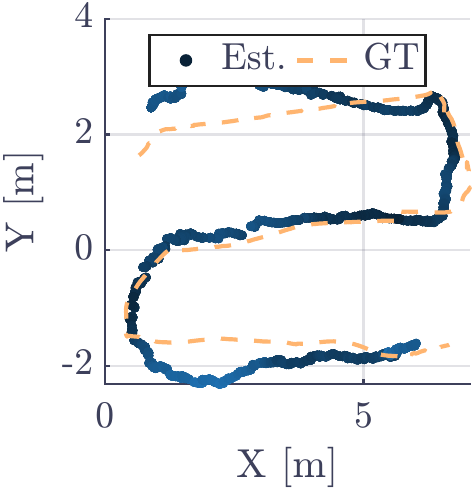}\label{fig:drone-grid:c}}
\\[-0.4ex]

\centering
\raisebox{7.5mm}[0pt][0pt]{%
\rotatebox{90}{%
{\color[HTML]{3D405B}\fontsize{7pt}{7pt}\selectfont Position [m]}}}
&
\subfloat[X position]{\includegraphics[width=0.29\linewidth]{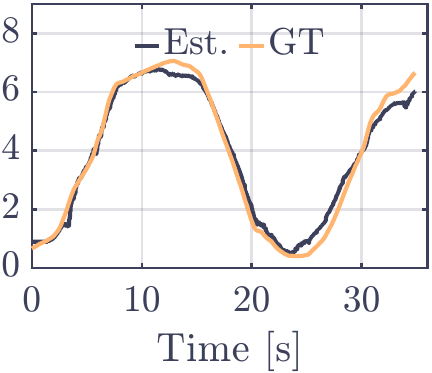}\label{fig:drone-grid:d}}
&
\subfloat[Y position]{\includegraphics[width=0.30\linewidth]{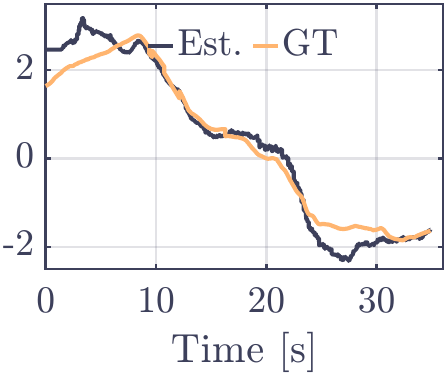}\label{fig:drone-grid:e}}
&
\subfloat[Z position]{\includegraphics[width=0.31\linewidth]{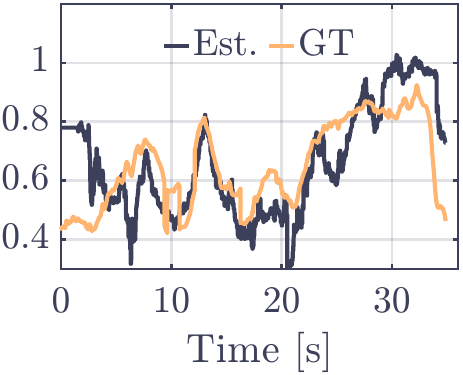}\label{fig:drone-grid:f}}
\\[-0.4ex]

\centering
\raisebox{6.5mm}[0pt][0pt]{%
\rotatebox{90}{%
{\color[HTML]{3D405B}\fontsize{6.5pt}{7pt}\selectfont Abs. Error [m]}}}
&
\subfloat[X abs. error]{\includegraphics[width=0.305\linewidth]{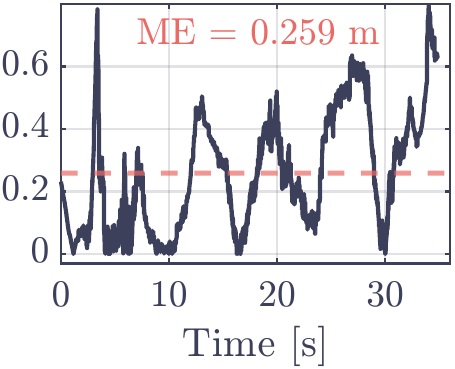}\label{fig:drone-grid:g}}
&
\subfloat[Y abs. error]{\includegraphics[width=0.305\linewidth]{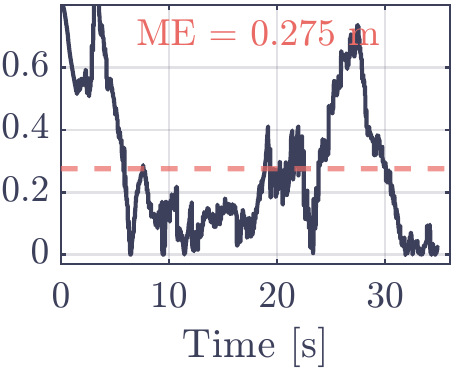}\label{fig:drone-grid:h}}
&
\subfloat[Z abs. error]{\includegraphics[width=0.305\linewidth]{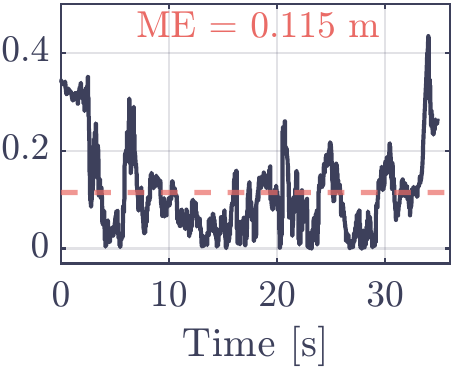}\label{fig:drone-grid:i}}

\end{tabular}
  \caption[ORB-SLAM3 trajectory and error analysis in dark scene with \ac{UAV}]{Experimental results in a dark indoor scene with \ac{UAV}. 
(a) \ac{UAV} with thermal cameras, 
(b) sample scene view, and (c) ground-truth trajectory. (d–f) Estimated trajectories along the X, Y, and Z axes from ORB-SLAM3 with T-RefNet are compared against ground truth. (g–i) Absolute position errors are shown for each axis, with ME values highlighted.}
  \label{fig:drone-grid}
\end{figure}

In the dark circular motion scenario in Fig.~\ref{fig:circle}, features cluster on one side of the scene, causing uneven keypoint coverage and degraded tracking. The X-axis shows the most significant deviations, with a mean error of approximately 0.20 m and frequent peaks of up to 0.70 m. Y and Z are more stable than X overall, with mean errors of about 0.12~m and 0.11~m; however, Y exhibits rare spikes up to 0.80~m, whereas Z's transients remain within about 0.50~m. These peaks are less frequent than in X, where fluctuations occur more consistently. Aggregated over all axes, the Euclidean mean error is about 0.26~m, highlighting the accuracy loss in circular motion under low light. Figure ~\ref{fig:thermal_rgb_comparison} further shows that ORB-SLAM3 with RGB input alone fails to maintain tracking in this scenario.

In the corridor experiment (Fig.~\ref{fig:drone-grid}), the \ac{UAV} flies through three different corridors: two in complete darkness and one with mixed illumination, where certain regions were lit while others remained in shadow. Within and at the end of each corridor, distinctive structures were placed to ensure the presence of detectable features. Additionally, aluminium foil strips were attached to selected surfaces to create low-emissivity targets, which appear as cold objects in thermal images even when they are at room temperature.

ORB-SLAM3 with T-RefNet enhanced thermal input successfully tracked the UAV's trajectory, achieving a Euclidean mean error of approximately 0.39 m across the three axes. This error reflects the increased difficulty posed by uneven lighting, cold-object distractors, and narrow passages. Nonetheless, the system maintained a continuous trajectory estimate, demonstrating robustness under mixed illumination and in the presence of thermally deceptive objects.

In summary, the three evaluation scenarios demonstrate that the proposed pipeline delivers stable localization across diverse conditions. In the bright feature-rich environments, thermal depth achieves accuracy comparable to RGB-D. In the dark circular motion, it sustains tracking with moderate accuracy loss where RGB completely fails. In \ac{UAV} corridor flights with mixed illumination and distractors, it maintains continuous trajectories within sub-0.4 m error. These results validate the robustness of the method under both handheld and \ac{UAV} setups in challenging environments.


\section{Conclusions and Future work}\label{sec:conclusion}



This paper presents a thermal-based depth estimation and SLAM framework for navigation in GPS-denied, low-light environments. A lightweight thermal-to-depth network with recurrent blocks, including \ac{RC}, was trained on radiometric VIVID++ and custom non-radiometric datasets, achieving state-of-the-art accuracy across both. Unlike prior methods that degrade on non-radiometric data, our recurrent design maintains temporal consistency under noise and low texture, requiring only $\sim$50k parameters versus $\sim$800k for ConvGRU. For localization, ORB-SLAM3 with T–RefNet enhanced thermal inputs yielded robust trajectories in both the handheld and the UAV experiments. In contrast, raw thermal or RGB inputs failed in darkness, demonstrating that the proposed pipeline extends SLAM to conditions where conventional vision breaks down.

 Despite its robustness, the framework can encounter challenges when the number of detected features is low, resulting in occasional tracking loss. Furthermore, artifacts inherent to thermal cameras, such as \ac{NUC}, may cause temporary intensity changes, disrupting feature stability and tracking. While real-time operation is feasible on embedded hardware for short sequences and moderate motion, prolonged or more dynamic scenarios still pose difficulties. As future work, we aim to mitigate these limitations by improving robustness against NUC artifacts, optimizing the pipeline for embedded platforms, and integrating depth estimation with obstacle avoidance modules to enable autonomous drone navigation and full trajectory estimation.



\bibliographystyle{IEEEtran}
\bibliography{References}

\end{document}